%% file: root_neurips_2023.tex
\documentclass{article}

\PassOptionsToPackage{numbers}{natbib}

\usepackage[final]{neurips_2023}

\usepackage[utf8]{inputenc} 
\usepackage[T1]{fontenc}    
\usepackage{hyperref}       
\usepackage{url}            
\usepackage{booktabs}       
\usepackage{amsfonts}       
\usepackage{nicefrac}       
\usepackage{microtype}      
\usepackage{xcolor}         

\usepackage{acronym}
\usepackage{amsmath,amssymb,amsfonts}
\usepackage{graphicx}
\usepackage{textcomp}

\usepackage{epsfig} 
\usepackage{mathptmx} 
\usepackage{times} 
\usepackage{bm}
\usepackage{algpseudocode}
\usepackage{caption}
\usepackage{subcaption}
\usepackage{booktabs}
\usepackage{tabularx}

\usepackage[para]{threeparttable}
\usepackage{multirow}
\usepackage{makecell}

\acrodef{ML}{machine learning}
\acrodef{RL}{reinforcement learning}
\acrodef{HRI}{human-robot interaction}
\acrodef{HHI}{human-human interaction}
\acrodef{ORCA}{Optimal Reciprocal Collision Avoidance}
\acrodef{IRL}{inverse reinforcement learning}
\acrodef{IL}{imitation learning}

\title{Improving Generalization in Reinforcement Learning Training Regimes for Social Robot Navigation}

\author{
    Adam Sigal$^{1}$,
    Hsiu-Chin Lin$^{2}$, and AJung Moon$^1$
\thanks{$^{1}$Adam Sigal and AJung Moon are with the Department of Electrical \& Computer Engineering, McGill University, Canada 
       {\tt\small adam.\{lastname\}@mail.mcgill.ca,ajung.\{lastname\}@mcgill.ca}
}
\thanks{$^{2}$Hsiu-Chin Lin is with the School of Computer Science and the
Department of Electrical \& Computer Engineering, McGill University, Canada
}
}

\begin{document}

\maketitle

\begin{abstract}

  \input{0-abstract}
\end{abstract}

\input{1-intro}
\input{2-background}

\input{3-method}
\input{4-experiments}
\input{5-conclusion}
\bibliography{ref}
\bibliographystyle{unsrt}

\end{document}

%% file: 0-abstract.tex
In order for autonomous mobile robots to navigate in human spaces, they must abide by our social norms. \Acf{RL} has emerged as an effective method to train sequential decision-making policies that are able to respect these norms.
However, a large portion of existing work in the field conducts both RL training and testing in simplistic environments.
This limits the generalization potential of these models to unseen environments, and the meaningfulness of their reported results. 
We propose a method to improve the generalization performance of RL social navigation methods using curriculum learning.
By employing multiple environment types and by modeling pedestrians using multiple dynamics models, we are able to progressively diversify and escalate difficulty in training.
Our results show that the use of curriculum learning in training can be used to achieve better generalization performance than previous training methods.
We also show that results presented in many existing state-of-the-art RL social navigation works do not evaluate their methods outside of their training environments, and thus do not reflect their policies' failure to adequately generalize to out-of-distribution scenarios.
In response, we validate our training approach on larger and more crowded testing environments than those used in training, allowing for more meaningful measurements of model performance.

%% file: 1-intro.tex
\section{INTRODUCTION}
\label{intro}
\noindent

Autonomous mobile robots are increasingly expanding into human spaces. 
In order for this process to continue smoothly and successfully, there are a number of social and technological factors that must be carefully considered.
These norms can vary depending on context and culture, and are often difficult to codify. 
For example, conceptions of personal space may be different for pedestrian traffic on the sidewalk versus an open space, indoors versus outdoors, speed of walking traffic, or time of day, among others. 

A myriad of approaches has been proposed as a means to create mobile robots that navigate in socially appropriate ways. While work exists which aims to explicitly codify human preferences when navigating around a robot \cite{mavro2022}, this remains a nascent field of research, and most social navigation approaches instead seek to emulate human behavior. One modern trend involves learning to replicate human-like behavior through interactive \ac{RL} training (i.e. where pedestrian agents can react to the robot agent, rather than using prerecorded data).
This is usually done in an environment where pedestrian agents follow predefined deterministic social navigation algorithms; most commonly either Social Force \cite{socialforce} or \ac{ORCA} \cite{orca}. 

Social Force explicitly models hypothetical pairwise attraction and repulsion forces between agents and obstacles. 
In \ac{ORCA}, each agent prunes its action space (permissible future velocities) at every time step, allowing for a mathematical guarantee of collision avoidance. 
Both of these models have intrinsic assumptions that lead to homogeneity of training data. 

Even in some of the most prominent \ac{RL} social navigation research, navigation models are trained in a single environment with fixed specifications, and on pedestrian navigation data generated by only one behavioral model \cite{chen, cadrl, lstm-rl}.

Conducting training using data exclusively generated by a single model of pedestrian behavior has been a point of criticism in the field~\cite{mavro}. 
A proof-of-concept experiment in this work illustrates the limitations of relying on a single deterministic model. The authors showed that a policy wherein an agent goes straight towards its goal, and stops whenever a pedestrian gets too close, outperforms \ac{ORCA}-based agents in time elapsed to reach the goal~\cite{mavro}. 
They posit that this indicates that training and testing exclusively on pedestrian data generated only by \ac{ORCA}, as many contemporary works do, is not a good indicator of performance in the real world.
Hence, in order to enable better social navigation behaviors of mobile robots, it is crucial that we scientifically examine the efficacy of current practices -- homogeneous modeling of pedestrians and simplicity of training and evaluation methods -- and challenge the status quo.


In this work, we explore how to improve the quantitative and qualitative performance of three state-of-the-art \ac{RL} social navigation models (CADRL \cite{cadrl}, LSTM-RL \cite{lstm-rl}, and SARL \cite{chen}) through training. Furthermore, we show that results presented in previous works are limited by the simplicity of their testing environments. In response, we present approaches to more meaningfully evaluate their generalizability. 
To this end, in this study we:
\begin{itemize}
    \item Conduct interactive RL training in an environment where some pedestrians are modeled by Social Force and some by \ac{ORCA}.
    \item Use a curriculum of different environment types during training.
    \item Conduct performance evaluation in unseen testing environments that are more challenging than those used in training, by increasing both crowd density and environment size.
\end{itemize}

The code used to conduct these experiments is publicly available at: \\
\href{https://github.com/RAISE-Lab/soc-nav-training}{\texttt{github.com/RAISE-Lab/soc-nav-training}}.

%% file: 2-background.tex
\section{BACKGROUND}
\label{background}

\subsection{Previous Work}
\label{prevwork}

In the early days of autonomous robot navigation, robots were designed to treat pedestrians as non-reactive obstacles, ignoring both human-human and human-robot reciprocity in navigation \cite{boren, dynamic, minerva, rhino}. 

Hence, these early attempts led robots to choose trajectories that humans found unexpected, encumbering the flow of traffic, and sometimes resulting in a \textit{reciprocal dance}~\cite{dance} -- where a robot would cause pedestrians to react suddenly and lead both agents to exhibit short vacillating motions before continuing onto their paths.

Later work sought to build robots that can navigate foot traffic while abiding by social norms in public spaces.
This included efforts to model the uncertainty of human trajectory predictions, and produced heuristics-based and data-driven models of pairwise classical \ac{HHI} within a crowd~\cite{socialforce, alahi2016social, vemula2018social}. 
Some prominent examples of this method are Thompson et al. \cite{thompson2009probabilistic} who proposed a probabilistic model of human path planning based on inference from real human data, Bennewitz et al. \cite{bennewitz} who built a hidden Markov model based on navigation in crowded spaces, and Unhelkar et al. \cite{unhelkar} whose motion prediction framework integrates biomoechanical features. 
%
%
However, using these models in navigational tasks remains a challenge due to the heavy computational cost involved, which can lead to the freezing robot problem \cite{unfreeze}.
More recent approaches to social navigation seek to couple the tasks of pedestrian prediction and motion planning either implicitly or explicitly. 
One example of this approach uses \ac{IRL} where social norms are inferred directly from human trajectory data through and contextualized into a reward function for \acf{RL} \cite{kretz, pineau, 
ziebart}. Another is \ac{IL}, a method that leverages example trajectories to learn socially-compliant navigation policies through conventional supervised learning \cite{tai2018, long2017deep, liu2018map}.
    \ac{IRL}-based approaches suffer from the \textit{reward ambiguity problem}, meaning that multiple different types of rewards could explain the same expert behavior \cite{ziebart2008maximum}. However, both \ac{IRL} and \ac{IL}, as well as more complex approaches such as those employing a combination of learning and control-based methods~\cite{nishi}, share further limitations; not only are they constrained by the amount of high-quality datasets available, but they also lack reactivity in training when using these datasets, as well as the degree to which they can be applied to new, unseen situations. 
    
    \Ac{RL} methods are able to bypass these issues by having agents learn in an interactive environment with a well-defined reward function. 
    One of the first prominent examples of this is CADRL \cite{cadrl}, which was conceived for two-agent pairwise collision avoidance. 
    This has limitations in more crowded settings, where all unique pairwise interactions must be modeled separately. This also means that, at a given time step, the agent's action must be chosen with an $argmax$ over the predicted state-action values, which quickly becomes computationally expensive. 
    Another notable approach is LSTM-RL \cite{lstm-rl}, which addresses this issue by considering all pedestrians together, ordered sequentially based on their proximity. 
More recently, SARL \cite{chen} proposed to model the importance of pedestrians in a more sophisticated manner, assigning each an \textit{attention score} using a self-attention mechanism. This method also employs \ac{IL} in the early phases of training to speed up learning, using expert demonstrations generated by \ac{ORCA}. 
In this way, the benefits of \ac{IL} are leveraged while avoiding the drawbacks;
agents get a head start at understanding navigation through IL, and subsequently do the majority of their learning through interactive \ac{RL}, where they are not limited to a static distribution of navigation behavior. 

\subsection{Social Force and ORCA}
\label{sf-and-orca}
\noindent
Social Force was one of the earliest pedestrian dynamics models to be successfully applied in autonomous robots in both simulation and the real world \cite{sud, ferrer, ferrer2}. It postulates that pedestrian path planning is based on attractive and repulsive forces from different points in space. These forces, which, as opposed to physical forces, are proportional to both velocity and acceleration, are grouped together into nonlinearly coupled Langevin equations as such:

\begin{equation}
    \vec{F}_i (t) = 
    \vec{F}^0_i
    + \sum_{j} \vec{F}_{i j}
    + \sum_{B} \vec{F}_{i B}
    + \sum_{k} \vec{F}_{i k}
\label{eq-sf}
\end{equation}
\noindent Where $\vec{F}_i (t)$ is the overall force exerted upon a pedestrian $i$ at time $t$, $\vec{F}^0_i$ is the attractive force of $i$ towards its goal, $\vec{F}_{i j}$ is the repulsive force between pedestrian $i$ and pedestrian $j$, $\vec{F}_{i B}$ is the repulsive force between pedestrian $i$ and a boundary $B$, and $\vec{F}_{i k}$ is the attractive force between pedestrian $i$ and a point of interest $k$. The full equations are given in \cite{socialforce}. While Social Force is useful for procedural trajectory generation, it is intrinsically limited in how it models pedestrian interaction, as it is based on hypothetical forces that do not truly exist in our world.

ORCA, on the other hand, works by collectively pruning possible actions from the action space of each agent $i$ at time $t$. In this setting, the action of the agent is its choice of velocity. After velocities that would lead to collision are pruned, the remaining set of possible actions is denoted ORCA$_i^t$. Under the assumption that all agents are following the same navigation policy, this results in a linear programming problem wherein collision avoidance is guaranteed, so long as it is mathematically possible. With this approach, the overall set of permitted velocities for a a given agent $i$ is:

\begin{equation}
    \text{ORCA}_i^t = D(\bm{0}, \bm{v}_{i, pref}) \cap \bigcap\limits_{i \neq j} \text{ORCA}_{i | j}^t \; \: \forall \text{ pedestrians } i,j 
\end{equation}
Where $D(\bm{0}, \bm{v}_{i, pref})$ denotes the open disc of velocities centered at $\bm{0}$ relative to human $i$ and of radius $\bm{v}_{i, pref}$ (preferred speed of $i$). 
$\text{ORCA}_{i | j}^t$ is the velocity closest to $\bm{v}_{i, pref}$ under the constraints of ORCA$_j^t$.
This also means that, when these constraints are violated, \ac{ORCA} can become susceptible to collisions and an overall decrease in performance \cite{mavro}.

\subsection{Reinforcement Learning for Social Navigation}
\label{probform}
\noindent
In this work, we consider a 2-dimensional robot agent navigating towards a goal point through a crowd of $n$ pedestrians. We approach this as a sequential decision-making problem using \ac{RL} as is done in previous notable \ac{RL} approaches \cite{chen, cadrl, sa-cadrl, lstm-rl}. 

The goal in a sequential decision-making \ac{RL} problem is to learn the optimal navigation policy $\pi^*$ and optimal value functions $V^*$ in order to reach the goal position $\bm{p}_g$. $\pi^*$ and $V^*$ are parameterized by a predefined reward function $R$ to shape the RL agent's behavior during training. In stochastic multiagent settings such as the present one, the calculation of $V^*$ is intractable, and therefore it must be approximated. There are different approaches to achieve this, for example using Deep V-Learning, as employed by \cite{chen, cadrl}.

We hypothesize that compared to earlier works~\cite{chen, cadrl, lstm-rl}, training on samples taken from diverse human navigation models will increase the robustness of the RL agents against unseen behavior.
In all of the algorithms tested in this study, most aspects of the respective RL training regimes are left unchanged.  
Furthermore, while previous works report exceptional results with their model architectures and training methods, reported tests are often conducted in an overly simplistic environment. In the example of SARL~\cite{chen}, evaluation was performed in simple environments of 4m radius circle with five pedestrians -- the same setting in which it was trained. In this work, we demonstrate that evaluating models in such simple environments is not sufficient to present meaningful results of model performance.

Other \ac{RL}-based social navigation models such as CADRL~\cite{cadrl} and LSTM-RL~\cite{lstm-rl} also share similar shortcomings.
As such, our work explores how to improve upon the shortcomings of RL methods such as these using new training and evaluation methods.

%% file: 3-method.tex
\section{METHODOLOGY}
\label{method}
\noindent 
We hypothesize that conducting training with a wider distribution of navigation scenarios and pedestrian behavior will increase robustness of RL models in novel and unseen situations. To this end, we evaluate the aforementioned three \ac{RL} models (SARL, CADRL, and LSTM-RL) in larger and more crowded environments with diverse pedestrian behavior, and thus oblige the models to perform over longer time horizons and in more challenging navigation scenarios. We also evaluate the performance of a simple \ac{ORCA}-based agent in the same scenarios for comparison against simpler, rule-based methods.

The \textit{baseline} training setting is the same as that which was used to train the 3 RL models in~\cite{chen}, where training occurs in the simple circle crossing environment (see Fig.~\ref{fig:traj-bl-cr},
Table~\ref{tab:env-types}) with pedestrians modeled only by ORCA.

We expect that allowing the \ac{RL} agent to learn from a more diverse experience set will allow it to better generalize to new situations. 
During training, we applied three new methods of pedestrian behavior diversification: 
\begin{enumerate}
    \item \textit{Diverse} setting: The same training process as~\cite{chen} is followed, but pedestrian behavioral models are randomized between ORCA and Social Force. In most cases, all obstacles are dynamic.
    
    \item \textit{Curriculum} setting: Since training takes place over 10,000 episodes in all of the \ac{RL} models considered, in the curriculum setting, training is split into two phases of 5000 iterations each. In the first, training takes place in the simple circle crossing environment, and in the second, training takes place in the simple square crossing environment (refer to Section \ref{simulator} for details). In the curriculum setting, some pedestrians become static obstacles, and some remain dynamic. We also tested the randomization of environment type at every episode, but found that performance was generally worse. 
    
    \item  {\em Curriculum+diverse} setting: The first phase of training remains the same (simple circle crossing, ORCA-only pedestrians). In the second phase, pedestrian behavior is randomized, as in the diverse setting, and training is conducted in the simple square crossing. Like in the \textit{curriculum} setting, here pedestrians become a mix of static and dynamic obstacles.
\end{enumerate}

\noindent 
We use a simple prefix to indicate the training method of a model. For example, CADRL agents with different training methods are expressed as follows: baseline: \textit{BL-}CADRL, diverse: \textit{D-}CADRL, curriculum: \textit{C-}CADRL, curriculum+diverse: \textit{CD-}CADRL.

Additionally, each agent $i$ has a radius $r_i$ and a preferred speed $v_{i, pref}$. 
In order to promote diversity in training, these attributes  are sampled uniformly for every human $i$ such that $r_i \sim U(0.3, 0.5)\ m$, and $v_{i, pref} \sim U(0.5, 1.5)\ m/s$.

Prior work conducts testing and training on the same small, simple environment~\cite{chen}. We hypothesize that testing done in larger and more crowded environments will be more representative of the trained model's true reasoning skills in new situations. 

To conduct experimentation, we adapted the codebase in~\cite{chen} and modified it to allow for a diverse set of navigational policies. In our experiments, we model pedestrian behavior using both ORCA and Social Force. For ORCA, we used the publicly available official implementation, RVO2~\cite{rvo}. Social Force was implemented by building upon work by Kreiss~\cite{kreiss}.
%
In the simulator used in the present study, each agent has a radius; however, Social Force models pedestrians as points. To account for this, as well as potential collisions under these circumstances, we define a new displacement value $\bm{d}_{i j}$ between two agents $i$ and $j$:
\begin{align}
    \bm{d}_{i j}' &= (\| \bm{p}_i - \bm{p}_j \|_2 - r_i - r_j) \odot \bm{u}_{i j} \\
    \bm{d}_{i j} &= \text{max}(\epsilon, \bm{d}_{i j}') 
\end{align}
\noindent Where $\bm{p}_i$ is the position of agent $i$, $r_i$ is its radius, $\bm{u}_{i j}$ is the unit displacement vector between $i$ and $j$, $\bm{d}_{i j}'$ is their radius-adjusted displacement vector, and $\epsilon$ is some real number very close to zero (e.g. $10^{-6}$). This is similar to agent radius accommodation done in existing Social Force variants~\cite{farina2017walking}, but avoids issues arising in collision cases which could otherwise lead to negative distance magnitudes.

%% file: 4-experiments.tex
\section{EXPERIMENTS}
\label{experiments}

\subsection{Implementation Details}
\label{implementation}
\noindent
To limit the effect of confounding variables, all model architectures, and all aspects of training except environments were kept consistent with their respective original specifications.

All experiments were conducted on a computer with an AMD Ryzen 9 5950X 16-Core Processor CPU, and an Nvidia GeForce RTX 3080 GPU. 
%
The training times of all studied RL models were found to be 
between $10 - 12$ hours. 

\subsection{Evaluation}
\label{eval}
\noindent 
The quantitative evaluation measures used in the present study are: success rate, collision rate, timeout rate, average time to goal (in case of success), rate of discomfort, and average closest distance. The average closest distance $d_T$ is the smallest distance the robot comes to any human over the time period $T$ of an epoch, averaged over all test epochs. 
The rate of discomfort is the proportion of total time the robot spends too close (less than 0.2 m) to a human agent over all test epochs. This metric is rooted in the theory of proxemics~\cite{prox}, a commonly used concept in social navigation research which defines quantitative measures of personal space in public settings~\cite{prox2, prox3}. 
As in previous work, a reinforcement learning episode can end in one of three ways: success (robot reaches goal), collision, or time-out \cite{cadrl, lstm-rl, chen}.

\subsection{Simulator}
\label{simulator}

\begin{figure}[t]
    \captionsetup{font=small}
    \captionsetup[sub]{font=small}
    \centering
    \begin{subfigure}{0.38\textwidth}
        \includegraphics[width=\linewidth]{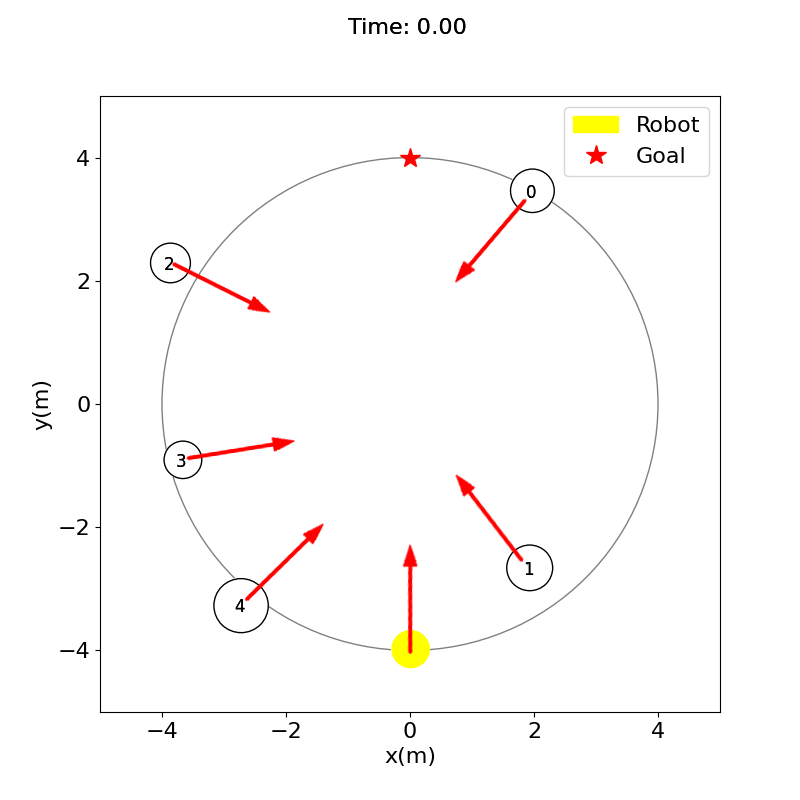}
        \caption{Circle Crossing}
        \label{fig:traj-bl-cr}
    \end{subfigure}
    \begin{subfigure}{0.38\textwidth}
        \includegraphics[width=\linewidth]{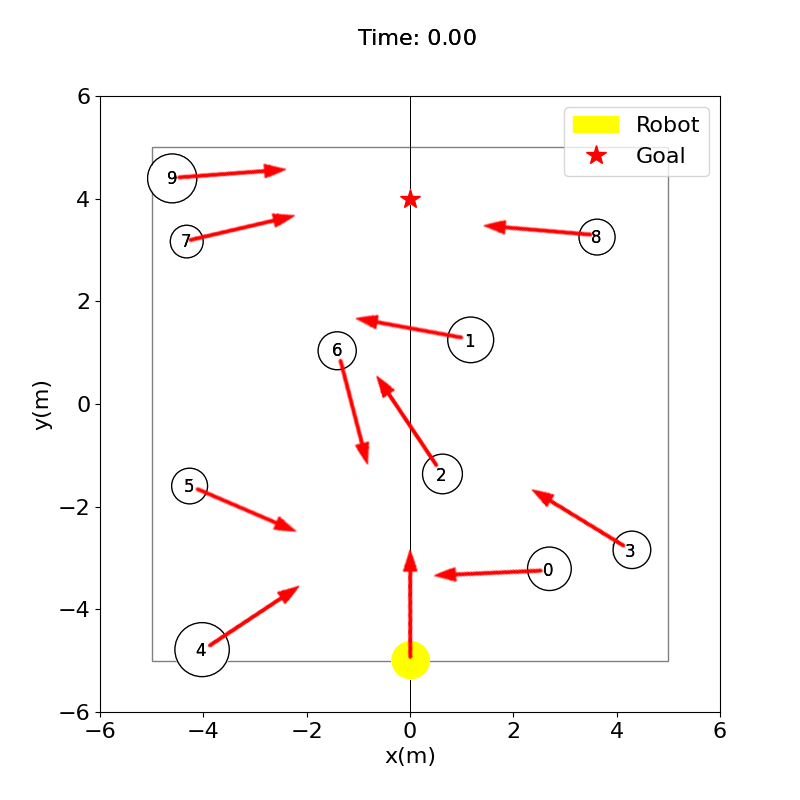}
        \caption{Square Crossing}
        \label{fig:traj-bl-sq}
    \end{subfigure}
    \caption{Examples of social navigation scenarios in the simple setting. White, numbered circles represent pedestrians, while the yellow circle represents the robot. The circular and squared outlines in the figures are not environment borders; they simply serve to illustrate the experimental environment type. All 6 environment types are described in Table \protect\ref{tab:env-types}.}
    \label{fig:trajectory}
\end{figure}

\noindent A simple top-down view simulator from~\cite{chen} built on OpenAI Gym is used in this study (Figs. \ref{fig:trajectory}, \ref{fig:attention}), where the action space of each agent is its 2-dimensional velocity. The observation space of an RL agent generally includes the positions and velocities of all pedestrians within a certain radius, but specific details depend on which model is used.
Experiments in the simulator were conducted on two main environment types: the ``circle crossing" and the ``square crossing" (Fig. \ref{fig:trajectory}).
In the circle crossing setting (Fig.~\ref{fig:traj-bl-cr}), humans start at a random position near the perimeter of the circle, and have a goal position roughly on the opposite side. This tends to create a social navigation challenge in the center of the circle which the robot agent must learn to manage efficiently. In the square crossing setting (Fig.~\ref{fig:traj-bl-sq}), humans have start points in a random position on either the left or right half of the square, and random goal points on the opposite side. This leads to a more diffuse social navigation task, with the added factor of having humans with goal points in the middle of the crowd becoming static obstacles once they reach them.

Each experimental environment is parameterized by two main variables: number of pedestrians $n$ and circle crossing radius $r$ (m) or square crossing width $w$ (m). Together, these give rise to a third characteristic metric: average crowd density $\rho_{\text{circle}} = \frac{n}{\pi r^2}$ or $\rho_{\text{square}} = \frac{n}{w^2}$ (humans / m$^2$). We conduct experiments with 2 density settings, a lower density (0.1 humans / m$^2$), and a higher density (0.2 humans / m$^2$). 

Table \ref{tab:env-types} presents the six environments used for training and testing. For each of the crossing types, three variants of density and size are used: simple, large, and dense. 

The \textit{simple} environment is of small size and lower density. 
The \textit{large} environment is lower density with an area that is roughly two times greater than that of the simple environment. Testing in this environment will illustrate the learned agent's capabilities to generalize to a longer-horizon task. 
The \textit{dense} environment keeps the same area as the simple environment, but approximately doubles the density of the crowd. Testing in this environment aims to demonstrate the RL agent's ability to navigate in more challenging settings where the risk of collision is far greater. Together, the four large and dense evaluation environments (referred to collectively as the \textit{Diverse-4} evaluation environments) move away from the toy settings in which previous models were trained and into more difficult and realistic scenarios. Performance in these environments is more indicative of navigation policy quality, as discussed further in Section \ref{sec:ablation}. 



\begin{table}[t]
\centering
\caption{Specifications of
environments used for training and testing. $n$ is the number of pedestrians, $r$ is the radius of a circle crossing (m), $w$ is the width of a square crossing (m), and $\rho$ is the mean crowd density (humans / m$^2$).}
\begin{tabular}{@{}cccccc@{}}
\toprule
\thead{Use}
& \thead{Env.}
& \thead{Crossing \\ Type}
& \thead{$n$}
& \thead{$r$ or $w$ \\ (m)}
& \thead{$\rho$ \\ (humans / m$^2$)}
\\ \toprule
\multirow{2}{*}{Training}  & \multirow{2}{*}{Simple}    & Circle        & 5   & 4          & 0.1    \\
        &     & Square        & 10  & 10         & 0.1    \\ \midrule
\multirow{4}{*}{\thead{Testing \\ \textit{(Diverse-4)}}}  & \multirow{2}{*}{Large}       & Circle        & 12  & 6          & 0.1    \\
        &        & Square        & 20  & 14         & 0.1    \\
        & \multirow{2}{*}{Dense}       & Circle        & 10  & 4          & 0.2    \\
        &        & Square        & 20  & 10         & 0.2    \\ \bottomrule
\end{tabular}
\label{tab:env-types}

\end{table}

CADRL, LSTM-RL, and SARL agents will be trained according to each of the 4 training settings described in Section \ref{method} (baseline, curriculum, diverse, and curriculum+diverse).




\input{4b-ablation.tex}

\input{4c-qualitative.tex}


%% file: 4b-ablation.tex
\subsection{Comparison of Evaluation Environments}
\label{sec:test-envs}

\begin{figure}[htbp]
  \centering
  \captionsetup{font=small}
  \includegraphics[width=0.45\textwidth]{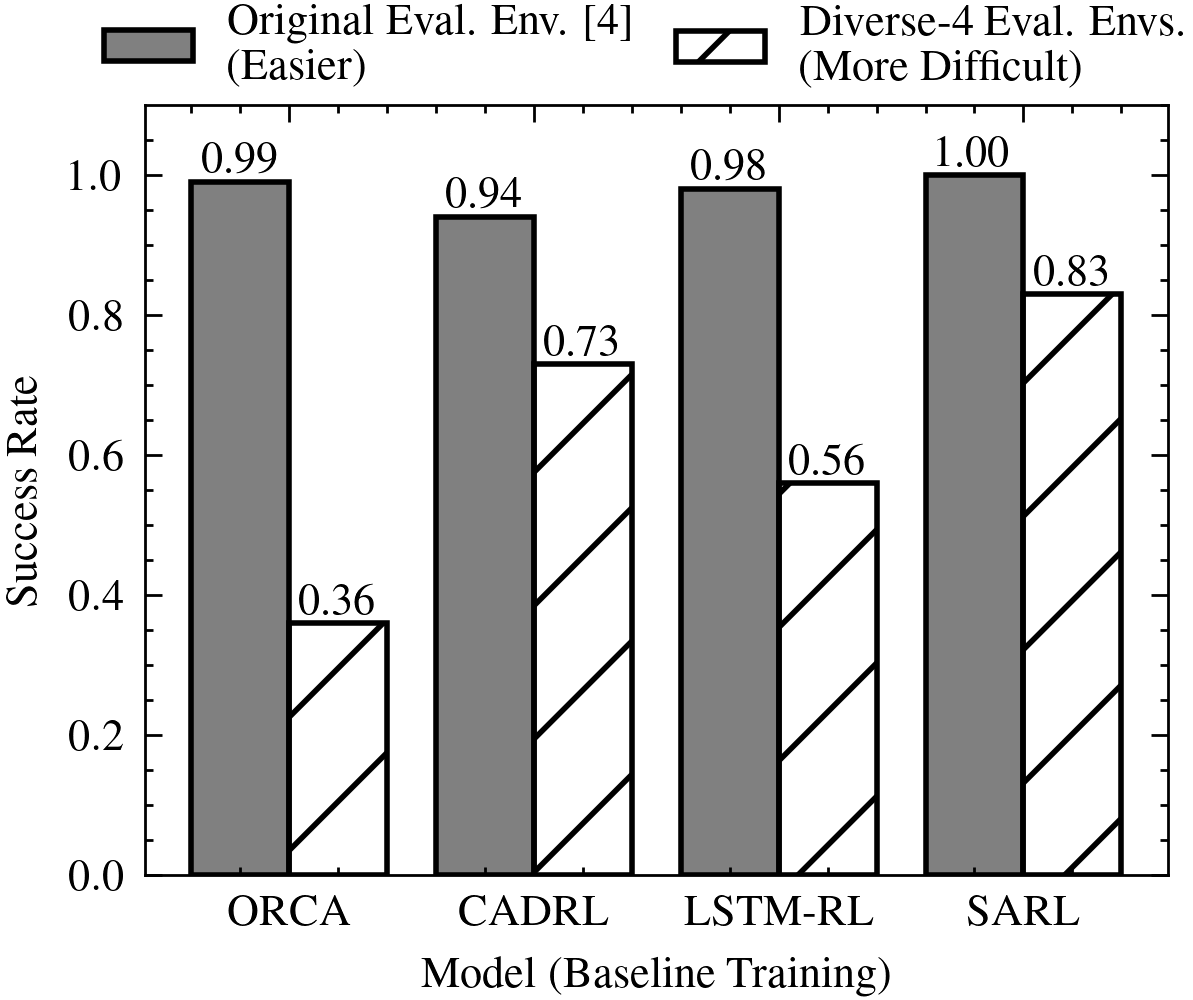}
  \caption{Evaluation environment comparison. Average success rate of baseline models (baseline training) on original evaluation environment (ORCA-only small circle crossing) vs. on the more challenging Diverse-4 evaluation environments. Our best method, CD-SARL, generalizes better to the Diverse-4 environments, with a 0.95 success rate (see Table~\ref{tab:ablat}).}
  \label{fig:test-comp}
\end{figure}


\noindent 

In Fig. \ref{fig:test-comp}, we compare model performance between the evaluation environment used in \cite{chen} (we call this the \textit{original} evaluation environment for clarity), and the more challenging Diverse-4 evaluation environments described in Section~\ref{simulator}. 
In our new, more challenging setting, all models with baseline training saw significant decreases in success rate. This demonstrates how limited the applicability of results from previous studies can be. More challenging and unseen evaluation settings, such as the Diverse-4 environments used in this study, are therefore crucial in providing meaningful results in RL social navigation experiments.

\subsection{Quantitative Ablation Study of Training Techniques}
\label{sec:ablation}

\begin{table*}[t]
	\centering
	 \caption{Ablation study of training methods described in Section \ref{implementation} (as well as ORCA). Presented is average performance over 50 episodes of each of the Diverse-4 testing environments: large and dense, circle and square crossings (± std. for applicable metrics). ($\downarrow$), ($\uparrow$) indicate whether higher or lower is better. When considering all metrics together, the \textbf{best overall performance (bold)} is achieved by CD-SARL
  (explanation in Section \ref{sec:ablation}).}
        \makebox[\textwidth][c]{
	\begin{tabular}{@{}cccccccc@{}}
		\toprule
        \thead{Training\\Methods} & Policy  & \thead{Success\\($\uparrow$)} & \thead{Collision\\($\downarrow$)} & \thead{Timeout\\($\downarrow$)} & \thead{Time (s)\\($\downarrow$)}      & \thead{Discomfort\\($\downarrow$)} & \thead{Closest Dist.\\$d_T$ (m) ($\uparrow$)}        \\  \midrule
		--             & ORCA    & 0.36    & 0.64       & 0.00    & 12.64 ± 2 & 0.24  & 0.10 ± 0.06 \\ \midrule
                             & CADRL   & 0.73    & 0.16      & 0.11    & 16.04 ± 4 & 0.10  & 0.13 ± 0.06 \\
		Baseline         & LSTM-RL & 0.56    & 0.38      & 0.06    & 15.75 ± 5 & 0.21  & 0.09 ± 0.06 \\
                             & SARL    & 0.83    & 0.17      & 0.00    & 11.20 ± 1 & 0.21  & 0.13 ± 0.05 \\ \midrule
                             & CADRL   & 0.77    & 0.21      & 0.02    & 13.51 ± 3 & 0.19  & 0.13 ± 0.06 \\
		Curriculum       & LSTM-RL & 0.25    & 0.38      & 0.37    & 21.94 ± 5 & 0.14  & 0.09 ± 0.06 \\
                             & SARL    & 0.80    & 0.07      & 0.13    & 14.61 ± 4 & 0.12  & 0.13 ± 0.05 \\ \midrule
                             & CADRL   & 0.73    & 0.12      & 0.15    & 17.46 ± 4 & 0.13  & 0.14 ± 0.05 \\
		Diverse          & LSTM-RL & 0.61    & 0.33      & 0.06    & 14.16 ± 4 & 0.19  & 0.10 ± 0.06 \\
                             & SARL    & 0.84    & 0.15      & 0.01    & 12.17 ± 1 & 0.18  & 0.13 ± 0.05 \\ \midrule
                             & CADRL   & 0.74    & 0.24      & 0.02    & 13.77 ± 3 & 0.22  & 0.13 ± 0.06 \\
		\textbf{Curr+Div}& LSTM-RL & 0.49    & 0.34      & 0.17    & 17.00 ± 5 & 0.18  & 0.10 ± 0.06 \\
                             & \textbf{SARL}    & 0.95    & 0.04      & 0.01    & 12.56 ± 2 & 0.13  & 0.14 ± 0.05 \\ \bottomrule
	\end{tabular}
        }
    \label{tab:ablat}
\end{table*}

An ablation study of the effects of the diverse and curriculum training settings was conducted for the CADRL, LSTM-RL, and SARL models. The results of this are presented in Table~\ref{tab:ablat}.
Results were collected for each of the three RL models over 200 test episodes for each training setting (50 episodes for each of the Diverse-4 evaluation environments).
These results are also compared with those of an ORCA-based robot agent. 

The time and closest distance to pedestrian ($d_T$) metrics present high levels of variance, both on agents trained with baseline methods, and with our proposed methods. This is to be expected in a highly stochastic multiagent navigation environment such as the one being studied, as the evolution of a crowd varies greatly between episodes.


In social navigation, performance metrics must always be looked at as a whole, since individual measurements can be misleading. For example, in Table~\ref{tab:ablat}, we see that ORCA has good performance in time to goal, but only at the expense of a high collision rate of 0.64, and high discomfort metrics. In this specific case, ORCA's poor performance arises because its assumptions of homogeneity are violated in the Diverse-4 testing environments. 
In fact, this is demonstrated most starkly in Fig.~\ref{fig:test-comp} where the ORCA agent goes from an almost-perfect success rate to one of just 0.36. This illustrates its fragility in non-homogeneous scenarios, and highlights the potential of \ac{RL}-based methods in contrast.


Overall, CADRL agents had fair, but not exceptional performance in the Diverse-4 testing environments, regardless of training method. CADRL's approach of evaluating pairwise interactions between the robot and each pedestrian independently is especially limiting in environments with more agents and higher density. This demonstrates the model's shortcomings in generalizing to more realistic situations. Since CADRL's performance did not significantly increase through the use of the new training techniques, it can be concluded that the model does not have the capacity to take advantage of a wider distribution of behaviour during training, further underlining its limited generalizability.

The new training methods show improvements in the performance of LSTM-RL as a whole when compared to baseline training, however, there is not a single method that performs better in all metrics. In the Diverse-4 testing environments, it performs worse than CADRL, which is in contradiction to the findings of previous work~\cite{chen}. Comparison between the original evaluation environment and the Diverse-4 environments shows that LSTM-RL has the greatest drop in success rate out of all tested RL models (Fig.~\ref{fig:test-comp}). In fact, LSTM-RL has poor performance overall, regardless of training method.  
This indicates that the approach of considering agents in a sequence ordered by proximity does not scale well to more challenging navigation scenarios.



The overall best performance was achieved using the Curriculum+Diverse training setting in CD-SARL. Its capacity to assimilate the diversity seen in this training setting may be due to SARL's unique approach of considering both interactions between pedestrians, as well as between pedestrians and the robot~\cite{chen}.
Its success rate is significantly higher than all other models, including SARL itself under other training methods. While the average closest distance to pedestrians is only slightly improved compared to BL-SARL, its frequency of causing discomfort is greatly reduced. This is a more important improvement as it is better to not enter into a pedestrian's personal space in the first place rather than to only minimize the degree of discomfort once that threshold has been crossed.
These findings demonstrate the usefulness of the proposed training techniques in achieving the best possible performance in social navigation tasks.

The combination of a marginally higher average time to goal and a lower discomfort rate -- as demonstrated by CD-SARL -- illustrates that the best social navigation policies will learn to manage the trade-off between efficiency and pedestrian discomfort to achieve an appropriate level of cautiousness.  

Conversely, the models with less ability to generalize to diverse and unseen testing environments (e.g. CADRL, LSTM-RL, and trivially, ORCA), were found to be less capable to learn from a wider distribution of behavior in training when given the chance.
Therefore, our findings suggest that the best \ac{RL} social navigation models have the capacity to assimilate patterns from a more diverse experience set, and use them to learn an appropriate balance of efficiency and cautiousness in navigation.

%% file: 4c-qualitative.tex
\subsection{Qualitative Experiments}



\begin{figure}[tb]
    \captionsetup{font=small}
    \captionsetup[sub]{font=small}
    \begin{subfigure}{0.3\textwidth}
        \includegraphics[width=\linewidth]{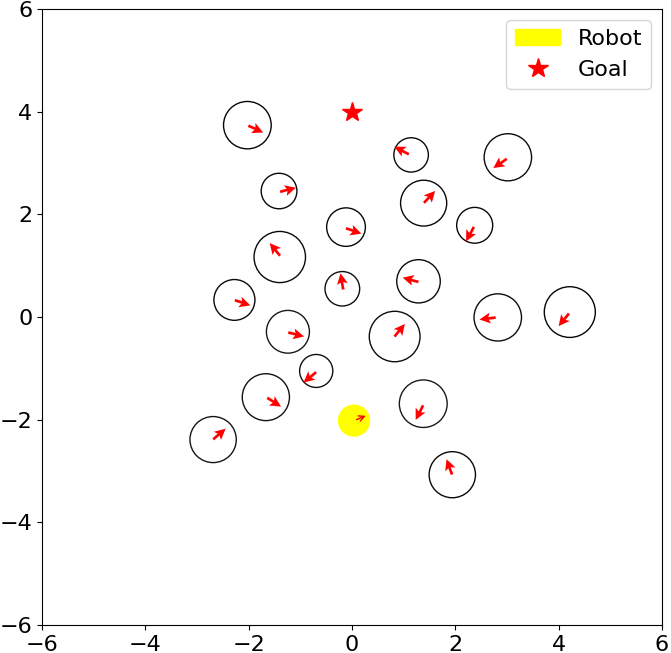}
        \caption{\protect\footnotesize Navigation Scene}
        \label{fig:attention}
    \end{subfigure}
    \hfill
    \begin{subfigure}{0.34\textwidth}
        \includegraphics[width=\linewidth]{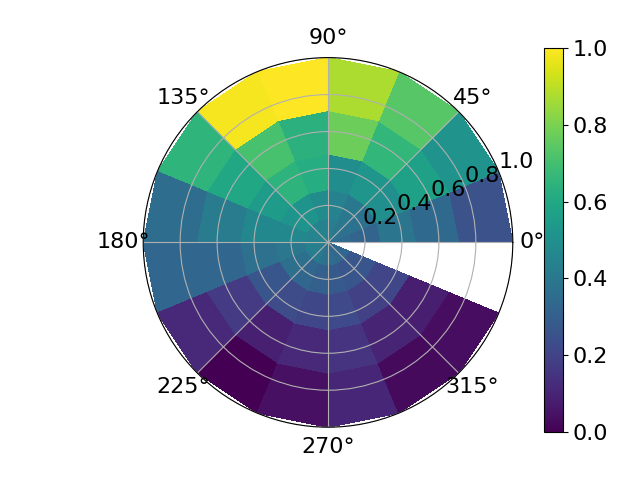}
        \caption{\protect\footnotesize BL-SARL Collision Case}
        \label{fig:coll-heat}
    \end{subfigure}
    \hfill
    \begin{subfigure}{0.34\textwidth}
        \includegraphics[width=\linewidth]{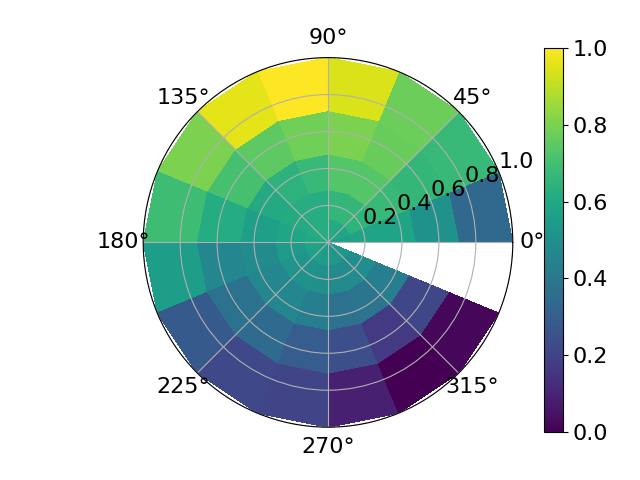}
        \caption{\protect\footnotesize CD-SARL Avoids Collision}
        \label{fig:no-coll-heat}
    \end{subfigure}

    \caption{(a) Scene in a diverse, dense square crossing after 2.25 seconds of navigation. Value estimations of each robot agent's action space in this scene (direction, and 5 increments of speed between 0 and $v_{pref}$) by BL-SARL (b) and CD-SARL (c). Values are shown on a gradient, where the highest estimated values are yellow, and the lowest are blue.}
    \label{fig:vf}
\end{figure}

As seen in Table \ref{tab:ablat}, BL-SARL has a collision rate almost five times greater than that of CD-SARL. Figs. \ref{fig:attention} and \ref{fig:vf} illustrate a scenario in which BL-SARL eventually collides with one of the pedestrians, while CD-SARL is able to successfully navigate to its goal. Videos of these two scenarios are available in the GitHub repository\footnote{\href{https://github.com/RAISE-Lab/soc-nav-training}{\texttt{github.com/RAISE-Lab/soc-nav-training}}}. 



These two models differ mainly in how they estimate the value of their possible future actions.
In Fig. \ref{fig:vf}, we can see that BL-SARL displays high confidence that it can continue at full speed through a dense part of the crowd. This eventually leads it to crash into a pedestrian 3 seconds later at $t=5.25$~s. On the other hand, CD-SARL has a more diffuse assignment of future state values, illustrating that it has learned to be more cautious than BL-SARL. It does this while still remaining reasonable and efficient, correctly identifying an area above itself that will soon become less crowded, and thus more easy to navigate through.

%% file: 5-conclusion.tex
\section{CONCLUSION}
\label{conclusion}

In this study, we empirically illustrate the limitations of training and testing social navigation \ac{RL} models in overly homogeneous environments. 
We addressed this by using curriculum learning and by diversifying pedestrian dynamics models during interactive RL training.  
Furthermore, our findings demonstrate that the results presented in many previous works do not adequately represent the generalizability of their models in unseen environments. 
In response, trained RL agents were tested in more challenging environments to more meaningfully evaluate generalizability.
We also find that the models that were best able to succeed in these novel settings were those that had the capacity to take advantage of the diversity of experience afforded by the new training methods. 

In future work, more intricate training curriculum -- e.g., including more pedestrian behavior models and environment types --  should be explored to further improve learning. Additionally, a natural extension of this work would be to validate its findings in a more complex simulator, and in a real-world experiment.